\documentclass[letterpaper, 10 pt, conference]{ieeeconf}  

\IEEEoverridecommandlockouts                              

\usepackage{subfigure}  
\usepackage{cite}  
\usepackage{graphicx}  
\usepackage{booktabs}  
\usepackage{amsmath}

\title{\LARGE \bf
VMTS: Vision-Assisted Teacher-Student Reinforcement Learning for Multi-Terrain Locomotion in Bipedal Robots
}

\author{Fu Chen, Rui Wan, Peidong Liu, Nanxing Zheng and Bo Zhou$^{*}$
\thanks{This work was supported by the National Natural Science Foundation (NnSF) of China under Grant 62073075. (Corresponding author: Bo Zhou.)}
\thanks{The authors are with the School of Automation, Southeast University, Nanjing 210096, China(e-mail: chenfu010728@gmail.com; wanrui@seu.edu.cn; 220242151@seu.edu.cn; 220221836@seu.edu.cn; zhoubo@seu.edu.cn;).}%
}

\begin{document}

\maketitle
\thispagestyle{empty}
\pagestyle{empty}

\begin{abstract}

Bipedal robots, due to their anthropomorphic design, offer substantial potential across various applications, yet their control is hindered by the complexity of their structure. Currently, most research focuses on proprioception-based methods, which lack the capability to overcome complex terrain. While visual perception is vital for operation in human-centric environments, its integration complicates control further. Recent reinforcement learning (RL) approaches have shown promise in enhancing legged robot locomotion, particularly with proprioception-based methods. However, terrain adaptability, especially for bipedal robots, remains a significant challenge, with most research focusing on flat-terrain scenarios.
In this paper, we introduce a novel mixture of experts teacher-student network RL strategy, which enhances the performance of teacher-student policies based on visual inputs through a simple yet effective approach. Our method combines terrain selection strategies with the teacher policy, resulting in superior performance compared to traditional models. Additionally, we introduce an alignment loss between the teacher and student networks, rather than enforcing strict similarity, to improve the student's ability to navigate diverse terrains. We validate our approach experimentally on the Limx Dynamic P1 bipedal robot, demonstrating its feasibility and robustness across multiple terrain types.


\end{abstract}

\section{INTRODUCTION}

Bipedal robots, owing to their anthropomorphic structure, offer significant potential and value across various application domains. However, the inherent complexity of their design presents a persistent challenge in control. Moreover, to effectively function within human-centric environments, visual perception plays a pivotal role. Yet, the integration of visual input further exacerbates the difficulty of the control task.

Over the past few decades, numerous researchers have proposed a variety of advanced methods in this field\cite{hornung2012adaptive,stumpf2014supervised,kanoulas2018footstep,rossini2023real}. Among the most prominent is Boston Dynamics' Atlas robot, which is capable of performing a wide range of complex movements\cite{mccrory2022humanoid,calvert2022fast,bostonDynamics2018parkourAtlas}. However, most of these methods rely on the integration of odometry to generate online maps and require substantial time for offline planning, often resulting in limited generalization performance.

\begin{figure}[t]
    \centering
    \includegraphics[width=0.5\textwidth]{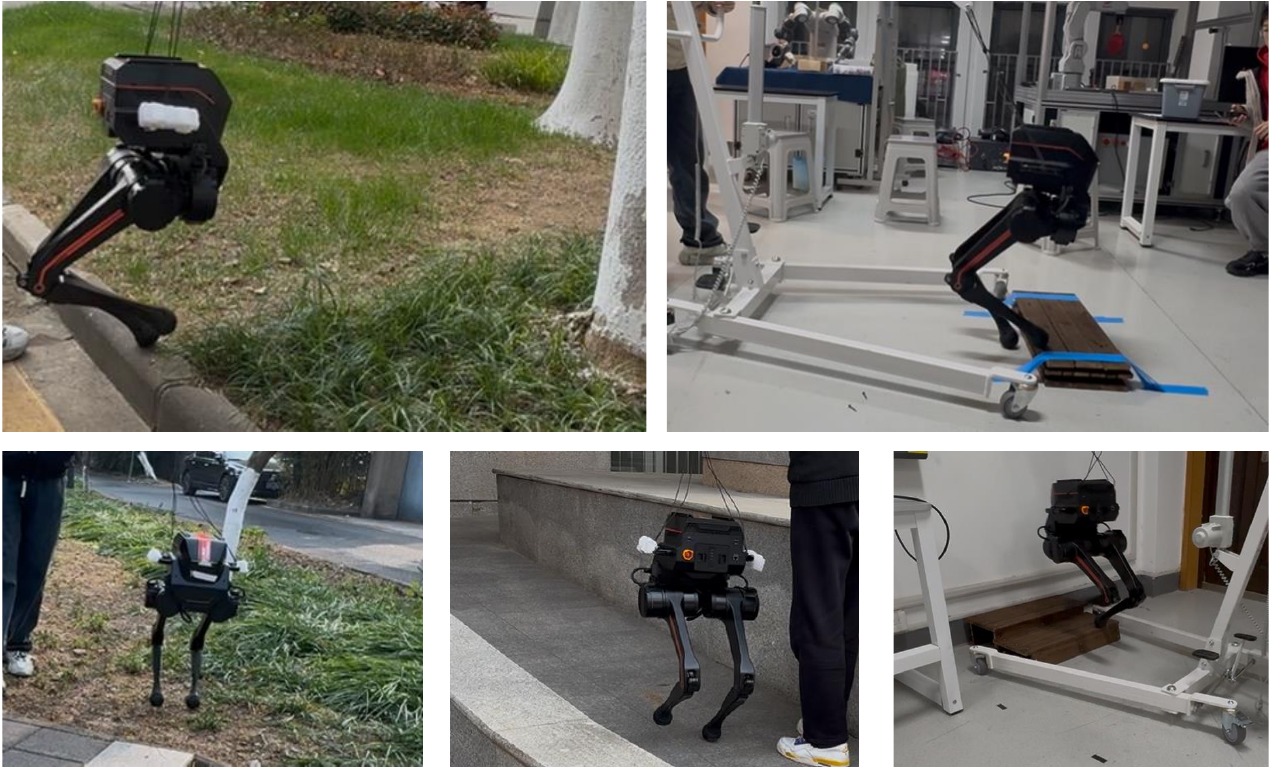}  
    \caption{We have developed an end-to-end vision-based reinforcement learning control strategy for bipedal robots, enabling them to overcome various terrains, such as 15 cm height differences, 30-degree slopes, and grasslands.}
    \label{fig:demo}
\end{figure}
Recently, reinforcement learning has yielded promising results in the control of legged robots, demonstrating significant advancements in the field\cite{lee2020learning,10161144,kumar2021rma,margolis2023walk,gu2024humanoid,rudin2022learning}. Currently, research on reinforcement learning for legged robots primarily focuses on proprioception. Lee et al. \cite{lee2020learning} were among the first to propose a teacher-student network learning approach on legged robot. In the teacher-student network, the teacher policy, endowed with privileged information, undergoes training in the first phase. During the second phase, the student policy, which relies solely on proprioceptive data, imitates the actions of the teacher policy. This methodology allows the student policy to achieve performance closely resembling that of the teacher policy while only utilizing the necessary information, making it a widely studied approach\cite{kumar2021rma,9981091,10375167}. Different to the teacher-student two-phase training approach, there are also numerous methods that require only a single-phase training process. 
DreamWaQ\cite{10161144} proposed the use of Variational Autoencoders (VAE) to estimate implicit representations of states. 
Wang et al.\cite{wang2024toward} applied this estimation method to bipedal robots and evaluated the impact of the estimated states on the locomotion performance of the robot. Fu et al.\cite{fu2023deep} proposed incorporating privileged learning of encoded proprioceptive information during the training process, and applied this approach to mobile manipulation tasks. Wang et al.\cite{ctswang} proposed a method for simultaneously training both teacher and student networks, which significantly enhances the performance of the student policy with minimal impact on the teacher policy. However, in the absence of external perception, overcoming terrain remains a significant challenge for legged robots, particularly bipedal robots. As a result, current research on bipedal robots primarily focuses on flat-terrain environments.

\begin{figure*}[t]
    \centering
    \includegraphics[width=\textwidth]{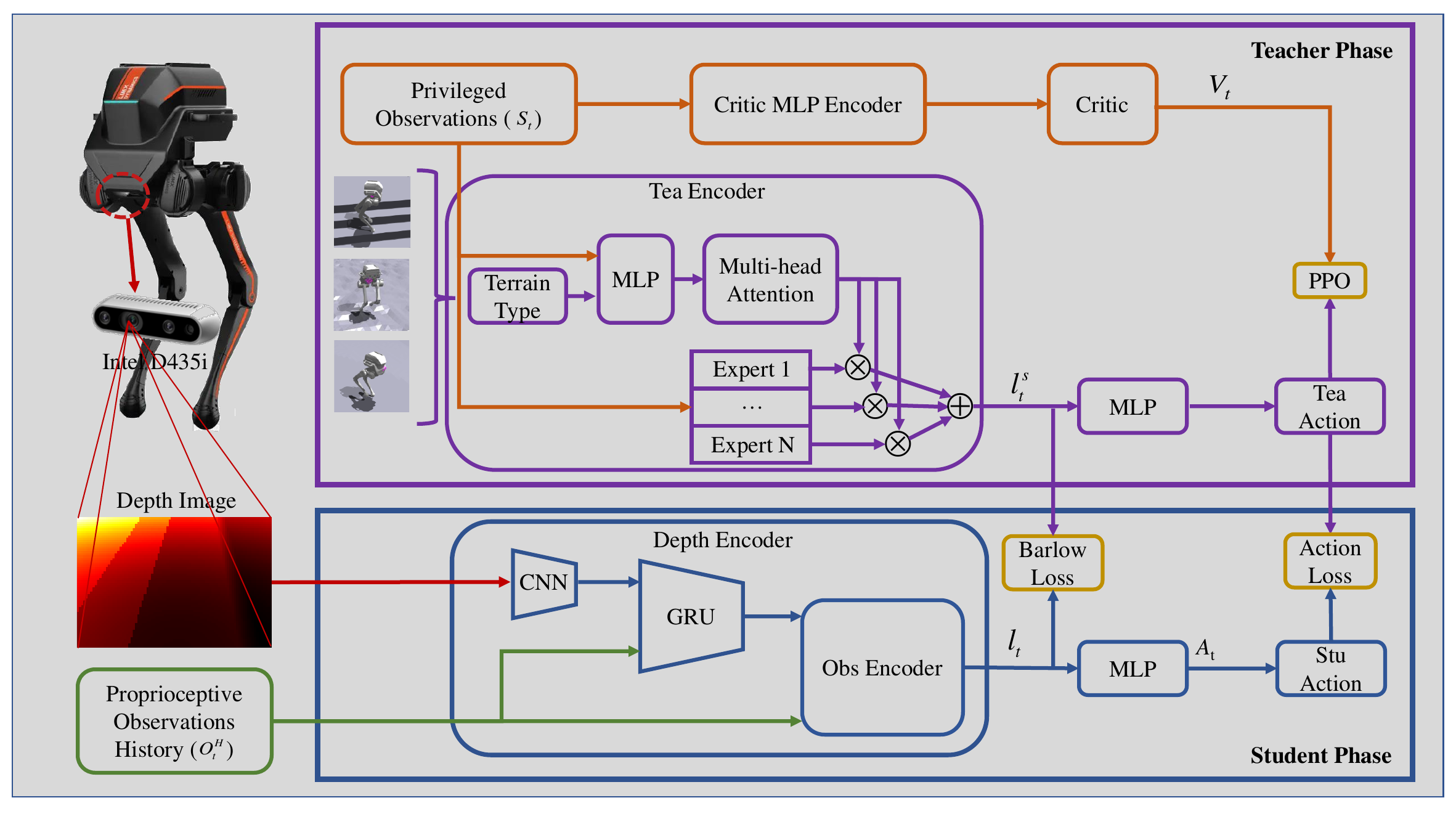}  
    \caption{Training Overview. In the first phase, the network inputs include proprioceptive information, the height map within a one-meter radius around the robot, and accurate physical parameters. Additionally, terrain classification and privileged information are utilized to adjust the weighting of the the mixture of expert, facilitating the training of a reliable teacher policy. In the second phase, the input information consists solely of the current depth images and proprioceptive data. The student policy is trained by aligning the encoded representations of both the student and teacher policies using Barlow Twins, while simultaneously employing supervised learning to guide the student policy to imitate the output of the teacher policy.}
    \label{fig:overview}
\end{figure*}

The incorporation of external perception aids in overcoming terrain challenges; however, it also increases the difficulty of training legged robots. Despite this, recent advancements have led to the development of several promising solutions\cite{ethterr,hoeller2024anymal,luo2024pie,yu2024walking,lai2024world,zhuang2023robot,cheng2024extreme,zhuang2024humanoid}. Reinforcement learning for legged robots with external perception is similarly categorized into teacher-student policies and single-phase training approaches. In the single-phase training framework, Luo et al.\cite{luo2024pie} extended DreamWaQ\cite{10161144} by incorporating depth images and employing a VAE-based method to implicitly estimate the surrounding height map of the robot. Meanwhile, Yu et al.\cite{yu2024walking} proposed a method that utilizes depth images and historical observations to explicitly estimate the height map and body state, enabling quadrupedal robots to overcome challenging terrains such as stairs, steep slopes, and large gaps. However, the single-phase training approach suffers from two major limitations: the significant difficulty of visual perception acquisition in Isaac Gym\cite{isaacgym} and the challenge of directly learning the environmental map. Consequently, current research on reinforcement learning for legged robots with external perception predominantly focuses on the teacher-student policy framework. \cite{ethterr} was among the first to propose using online-generated height maps as external input for enabling the student policy to imitate the teacher policy's behavior, achieving favorable experimental results. \cite{zhuang2023robot} introduced the direct use of depth images as external input, eliminating the need for online height map generation. \cite{cheng2024extreme} employed waypoints as tracking targets to enhance locomotion performance. \cite{zhuang2024humanoid} applied parkour techniques to humanoid robots, successfully enabling them to traverse higher platforms and gaps. 

However, the limitation of this two-phase training paradigm lies in the fact that the student policy cannot fully imitate the teacher policy's behavior due to discrepancies in the input information, which may even affect the student policy itself. Moreover, since most current training methods treat different terrains as a single task and employ the same policy network, they overlook potential conflicts that may arise under varying terrain conditions, potentially leading to a degradation in performance.

In this paper, we propose a novel the mixture of experts teacher-student reinforcement learning strategy, which improves the performance of teacher-student policies based on visual inputs through a simple approach. Inspired by\cite{lan2024contrastive} and\cite{zbontar2021barlow}, we introduce a mixture of experts model alongside Barlow Twins. By integrating terrain selection strategies with the teacher policy, our method outperforms traditional teacher-student models. Additionally, our approach focuses on aligning the input information between the teacher and student, rather than forcing them to be identical, thereby improving the student's ability to navigate diverse terrains. We deployed our algorithm on the Limx Dynamic P1 bipedal robot and tested it across a variety of terrains.

In summary, the contributions of this paper are as follows:
\begin{itemize}
\item First, our framework integrates a mixture of expert models with teacher-student policy learning, where the teacher strategy adaptively adjusts its policy based on terrain, providing a novel vision-aided reinforcement learning framework for legged robots. 
\item Second, compared to previous approaches that directly distill the teacher's policy to the student, we introduce an implicit alignment and redundancy reduction in the observation feature space between the teacher and student prior to distillation, thereby enhancing the student's performance. 
\item Finally, we propose a smooth foot-trajectory tracking reward function to mitigate the sim-to-real gap, with comprehensive simulation and physical experiments validating the overall efficacy and superiority of our approach.
\end{itemize}

\section{Method}

\subsection{Overview}
As shown in Fig. \ref{fig:overview}, our proposed framework consists of three main components: the teacher policy, the student policy, and the mixture of experts model. Due to the high resource consumption and difficulty in effectively learning visual features in single-phase training, we have opted for a two-phase teacher-student strategy. The input to the student policy includes only the data required for policy deployment, whereas the teacher policy's input, in addition to proprioceptive information, incorporates other privileged information available from the simulation environment. The teacher policy is optimized using the Proximal Policy Optimization (PPO) algorithm\cite{schulman2017proximal}. Additionally, the mixture of experts model (detailed in Section \ref{sec:Experts}), which has been shown to enhance performance in multi-task learning, is integrated into the teacher policy to improve performance across multiple terrains. During the student policy's training process, Barlow Twins (explained in Section \ref{sec:Twins}) is employed to align the encoded feature space, while the student policy is forced to imitate the teacher policy's actions, enabling it to approximate the teacher policy’s behavior.
\subsubsection{Teacher Policy}
The input to the teacher policy $s_t$, includes the proprioceptive observation $o_t$, the surrounding height map $m_t$, and the physical parameters $o_p$ obtained from the simulator, such as mass, velocity, and friction coefficients. 
\begin{equation}
    s_t = <o_t,o_p,m_t>
\end{equation}
The proprioceptive observation $o_t$ is a 29-dimensional vector, comprising the angular velocity $w_t$ and gravity vector $g_t$ directly obtained from the IMU, joint positions $\theta_t$, joint velocities $\dot{\theta}_t$, the action $a_t$ from the previous time step, user commands $c_t$, and the reference contact phase $p_t$ of the foot.
\begin{equation}
    o_t = <w_t, g_t, \theta_t,\dot{\theta}_t,a_t,c_t,p_t>
\end{equation}
Here, $p_t$ represents the reference contact phase of the foot, introduced based on \cite{gu2024humanoid} as a square wave function with a period of 0.5s, designed to address the balancing challenge in footed robots that lack a sole.

\subsubsection{Student Policy}
The input to the student policy includes the historical proprioceptive observations $o_t^H$ and the depth information $d_t$ processed through convolutional and recurrent neural networks.

\subsubsection{Value Network}
The input $s_t$ to the evaluation network is the same as that of the teacher network. Therefore, the loss function for the teacher policy training is:
\begin{equation}
    \mathcal{L} = \mathcal{L}^{ppo,s} + \mathcal{L}^{value}
\end{equation}

\subsubsection{Action Space}

The output action $a_t$, is a 6-dimensional vector, corresponding to the 6 joints of the bipedal robot, representing the deviation between the target positions and the default joint positions. Accordingly, the PD control for the joints is defined as follows:
\begin{equation}
    \tau = k_p\times(a_t-\theta_{d}) - k_d\times\dot{\theta}
\end{equation}

\subsection{Mixture of Experts}
\label{sec:Experts}
\begin{figure}[htbp] 
	\centering  
	\vspace{10pt} 
	\subfigtopskip=8pt 
	\subfigbottomskip=8pt 
	\subfigcapskip=3pt 
	\subfigure[stair]{
		\label{level.sub.1}
		\includegraphics[width=0.4\linewidth]{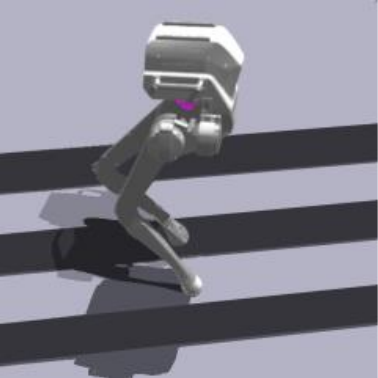}}
	\quad 
	\subfigure[slope]{
		\label{level.sub.2}
		\includegraphics[width=0.4\linewidth]{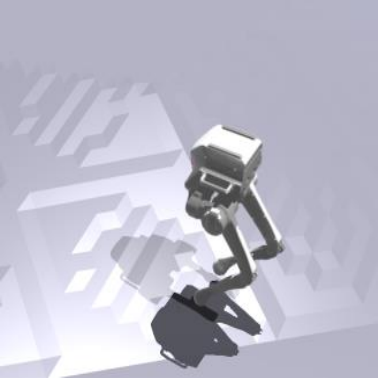}}
	\subfigure[rough]{
		\label{level.sub.3}
		\includegraphics[width=0.4\linewidth]{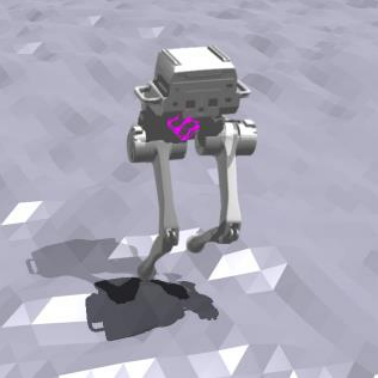}}
	\quad
	\subfigure[plane]{
		\label{level.sub.4}
		\includegraphics[width=0.4\linewidth]{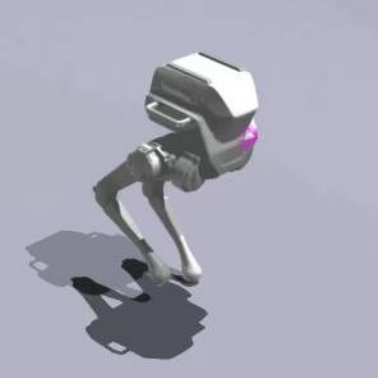}}
	\caption{Terrain type}
	\label{fig:terrain}
\end{figure}

To achieve locomotion control across various terrains, we adopted a curriculum training approach similar to that in \cite{rudin2022learning}, categorizing terrains into several types, such as flat ground, slopes, rough terrain, and stairs, as shown in Fig. \ref{fig:terrain}. Each terrain type progressively increases in difficulty as the robot’s performance improves during training, such as by increasing the slope angle, staircase height, or the roughness of the ground surface. However, training on different terrains may affect the performance of the policy across these terrain types. Inspired by \cite{lan2024contrastive}, we introduced a mixture of experts model during the training of the teacher policy. We select the number of experts to be 4. The encoder for the terrain categories and observation inputs is a three-layer fully connected network. After applying multi-head attention and Softmax, the outputs are multiplied by the expert mixture model to obtain the final teacher policy encoding. Therefore, the selection mechanism of the mixture of experts is represented as follows:
\begin{equation}
    a_1,a_2,a_3,a_4 = Softmax(W(g(s_t,t_p)))
\end{equation}
Here, $t_p$ represents the terrain category, $a_i$ denotes the expert weights, $W$ represents the attention, and $g$ refers to the fully connected network.

\subsection{Barlow Twins}
\label{sec:Twins}
During the student policy training phase, as proposed in \cite{ethterr}, the output of the teacher policy supervises the output of the student policy, while the privileged information from the teacher policy further supervises the encoded information of the student policy to facilitate the imitation of the teacher's behavior. We adopt a similar approach; however, instead of directly forcing the student’s encoded information to match the teacher's, we utilize the Barlow Twins\cite{zbontar2021barlow} method to align the feature spaces of the student and teacher encodings, rather than enforcing their exact proximity.
Therefore, the loss function for the student training process can be expressed as:
\begin{equation}
    \mathcal{L} = MSE(\hat{a_t},a_t) + \sum_i(1-C_{ii})^2 + \lambda\sum_i\sum_{i\neq j}(1-C_{ij})^2
\end{equation}
Here, $\hat{a_t}$ represents the output of the teacher policy,  $\hat{a_t}$ denotes the output of the student policy, and $C$ refers to the cross-correlation matrix between the student encoding $Z_a$ and the teacher encoding $Z_b$.

\section{Train Details}
\subsection{Reward Function Design}
To achieve stable locomotion for the bipedal robot, we designed the reward function as shown in Table \ref{tab:reward}, with reference to \cite{ctswang}. Here, $f_{feet}$ represents the foot contact force, $h$ denotes the body height, $t_{air}$ indicates the foot airtime, $p_{left}$ and $p_{right}$ represent the positions of the left and right feet in the body coordinate system, $h_{feet}^{max}$ refers to the maximum height of feet lift, $n_{collision}$  denotes the number of collision points, $\tau$ represents the joint torques, and $q_{hip}$ indicates the position of the hip joint.
\begin{table}[t]
\centering
\caption{Reward Function Elements}
\begin{tabular}{ccc}
\toprule  
Reward & Equation & Weight \\
\midrule  
Lin. velocity tracking  & $exp(-4||v_{xy}^{cmd}-v_{xy}||^2)$   & 7.5   \\
Ang. velocity tracking  & $exp(-4||w_{z}^{cmd}-w_{z}||^2)$  & 4.0   \\
Lin. velocity (z)  & $v_z^2$   & -0.5   \\
Ang. velocity (xy) & $w_{xy}^2$ & -0.06\\
Orientation & $g^2$ & -6.0\\
Base Height & $||h^{target}-h||$ & -10.0\\
Joint acceleration & $\ddot{q}^2$ & -$2.5\times10^{-7}$\\
Action rate & $||a_t-a_{t-1}||_2^2$ & -0.01 \\
Feet air time & $||max(t_{air}, t_{min})||$ & 60 \\
Joint torque & $||\tau||_2^2$ & -$2.5\times10^{-5}$ \\
Feet contact force & $||f_{feet}-f_{max}||_2^2$ & -0.01 \\
Feet distance & $||p_{left}-p_{right}||_2^2$ & -100 \\
Unbalance feet height & $||h_{left}^{max}-h_{right}^{max}||$ & -60 \\
Unbalance feet air time & $||t_{left}^{max}-t_{right}^{max}||$ & -300 \\
Collision & $n_{collisin}||$ & -30 \\
Joint torque limitation & $max(\tau-\tau_{limit},0)$ & -0.1 \\
Target feet height & $||h_{feet}^{target}-h_{feet}||$ & -6.0 \\
Feet position (x) & $||(p_x^{left}-p_x^{right})||$ & -3.0 \\
Hip Position & $||q_{hip}-q_{hip}^{default}||$ & -1.5 \\
Feet Contact Number & $r^{fn}$ & 2.5 \\
Lin. velocity smooth & $||\dot{v}_{xy}^t-\dot{v}_{xy}^{t-1}||$ & -0.05 \\
Survival & $t_{survival}$ & -0.05 \\
Feet Velocity & $r^{fv}$ & -0.5 \\

\bottomrule  
\end{tabular}

\label{tab:reward}
\end{table}
In order to facilitate smooth foot landing during the training process, we impose a penalty on the foot contact force. However, due to the inherent inaccuracies in the foot contact force measurements within the simulator, we supplement the force penalty with an additional penalty on the vertical velocity of the foot when it is within a certain distance from the ground, in order to achieve the same objective. The expression for this penalty is as follows:
\begin{equation}
    r^{fv} = v_z^{feet}\times(h_{feet} < 0.05)
\end{equation}

Due to the lack of a footplate structure in point-foot bipedal robots, maintaining balance becomes challenging. Therefore, we introduce the foot reference phase as described in \cite{gu2024humanoid}, and penalize the robot's foot phase when it deviates from the reference phase. The reward formula is as follows:
\begin{equation}
    r^{fn} = \sum_i^N(1\times(C_i=S_i)+(-1)\times(C_i\neq S_i))
\end{equation}
\begin{equation}
    C_i = \begin{cases}
0  & \text{ if } sin(\frac{2\pi t}{T}+\theta) \ge 0 \\
1  & \text{ if } sin(\frac{2\pi t}{T}+\theta) < 0
\end{cases}
\end{equation}
\begin{equation}
    \theta = \begin{cases}
0  & \text{ if } i=left \\
\pi  & \text{ if } i=right
\end{cases}
\end{equation}
Here, $N$ represents the number of feet, $C_i$ denotes the reference phase for the i-th foot, and $S_i$ represents the actual phase of the i-th foot, T=0.5s denotes the period.

Due to the tendency of bipedal robots to continuously raise the foot height during training in an attempt to traverse various terrains, this behavior is detrimental to maintaining balance. Therefore, to constrain the foot height during training and ensure the smoothness of the foot trajectory, we have designed the following reference foot trajectory:
\begin{equation}
    h_{feet,i}^{target} = \frac{1}{2}max(h_{feet}^{max}\times(1-cos(\frac{4\pi t}{T}), 0)\times \neg C_i 
\end{equation}
Here, $h_{feet}^{max}$ represents the maximum foot height.
\begin{table}[t]
\centering
\caption{Parameters Setup}
\begin{tabular}{ccc}
\toprule  
Parameter & Value & Unit \\
\midrule  
Base height  & 0.66  & m  \\
Feet height  & 0.03  & m  \\
Action scale  & 0.25  & -  \\
Decimation  & 4  & -  \\
Kp  & 40.0  & $N/rad$  \\
Kd  & 2.0  & $N\cdot s/rad$  \\
Lin. velocity range (x)  & [-0.5, 0.5]  & $m/s$  \\
Lin. velocity range (y)  & [-0.2, 0.2]  & $m/s$  \\
Ang. velocity range (y)  & [-0.5, 0.5]  & $rad/s$  \\
\bottomrule  
\end{tabular}
\label{tab:param}
\end{table}
\subsection{Environment Setup}
\label{subsec:environ}
We employed NVIDIA Isaac Gym\cite{isaacgym} as the simulation environment to train 8192 robots in parallel, with both the teacher and student policies utilizing 8192 robots. The teacher policy converged after approximately 4000 training episodes, requiring around 3 hours on an RTX 3090 GPU. During the training of the student policy, the substantial memory consumption and longer acquisition times for visual information in Isaac Gym significantly increased the training time. Specifically, training the student policy for 4000 episodes, building on the teacher policy, took approximately 120 hours. To address this, we utilized NVIDIA Warp\cite{warp2022} to store terrain mesh data during terrain initialization. When acquiring depth images, we adopted a ray-casting approach based on the robot's camera pose and physical properties, rather than relying solely on Isaac Gym's visual data acquisition. This modification, leveraging NVIDIA Warp for depth images retrieval, reduced the training time for the student policy to approximately 16 hours.During the training process, certain training parameters of the robot, such as PD parameters, target base height, foot end height, and others, are configured as shown in Table \ref{tab:param}.
\begin{table}[t]
\centering
\caption{Domain Randomization}
\begin{tabular}{ccc}
\toprule  
Parameter & Range & Unit \\
\midrule  
Payload mass  & [-1, 2]  & Kg  \\
Center of mass shift & [-3,3]$\times$[-2,2]$\times$[-3,3]  & cm  \\
$K_p$ Factor & [0.8,1.2]  & N/rad  \\
$K_d$ Factor & [0.8,1.2]  & N$\cdot$s/rad  \\
Friction & [0.2,1.6]  & -  \\
Restitution & [0.0,1.0]  & -  \\
Step delay & [0, 15]  & ms  \\
Push vel (xy) & 0.5  & m/s  \\
\midrule  
Camera position & [-5,5]$\times$[-10,10]  & mm  \\
Camera angle (pitch) & [-1,1]  & deg  \\
Camera Fov & [86,88]  & deg  \\
\bottomrule  
\end{tabular}
\label{tab:domain}
\end{table}
\subsection{Domain Randomization}
To enhance the model's generalization ability and reduce the simulation-to-reality gap, thereby facilitating the deployment of models trained in simulation on real robots, we applied threshold randomization within a specified range to certain physical parameters of the robot in the simulation. These parameters include PD controller gains, friction coefficients, joint masses, camera parameters, and others. The specific settings for the threshold randomization are presented in Table \ref{tab:domain}.

\section{EXPERIMENT}
\subsection{Simulation Experiments}
\begin{figure}[htbp]
    \centering
    \includegraphics[width=0.5\textwidth]{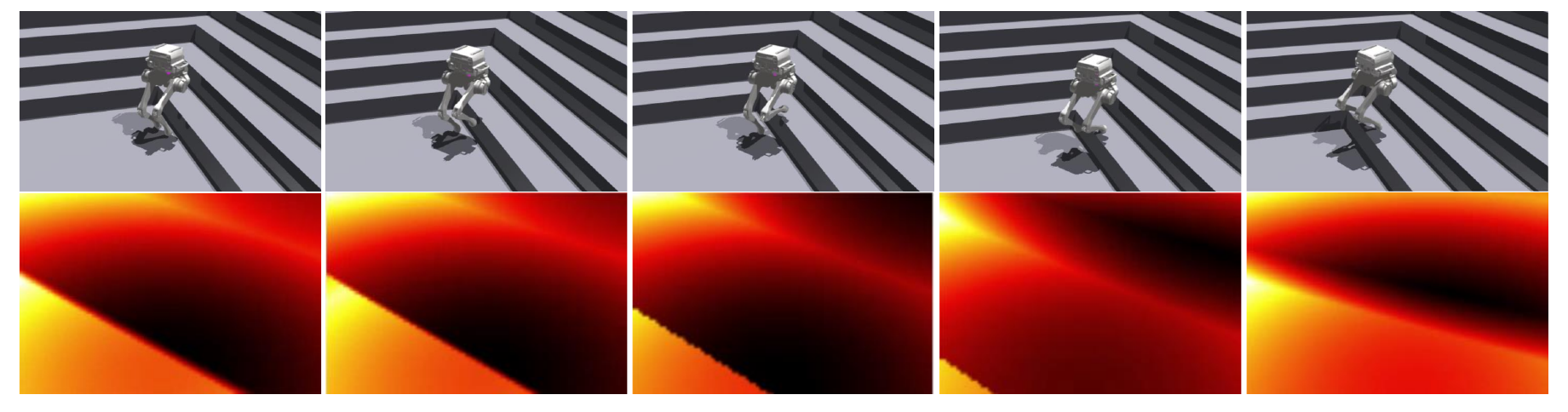}  
    \caption{The motion timing diagram of a bipedal robot with visual assistance in simulation. The lower section of the diagram illustrates the depth images obtained through NVIDIA Warp and ray casting during the robot's movement.}
    \label{fig:sim}
\end{figure}
To evaluate the effectiveness of the proposed method, we performed a comparison with several alternative approaches within the simulation environment. All methods were trained under the identical setup described in Section \ref{subsec:environ}, with the reward function held constant. The comparison was conducted by only modifying the network architecture and loss functions. The detailed descriptions of these methods are as follows:
\begin{itemize}
    \item Blind: The input consists solely of proprioceptive information, and training is conducted using the Proximal Policy Optimization (PPO) algorithm.
    \item Teacher-Student policy(TS): The two-phase teacher-student strategy utilizes depth images and proprioceptive information as inputs. The teacher policy is trained first, followed by the training of the student policy. A detailed description of this approach can be found in \cite{cheng2024extreme}.
    \item PIE: The single-phase strategy simultaneously trains the model while explicitly or implicitly predicting information such as velocity, body height, and height maps, as referenced in \cite{luo2024pie}.
\end{itemize}
To ensure a fair comparison, for the single-phase training method, we selected data from a total of 8000 iterations. For the two-phase training method, we chose 4000 iterations from each phase, resulting in a total of 8000 iterations, with 4000 iterations from phase one and 4000 iterations from phase two.

\begin{figure}[htbp]
    \centering
    \includegraphics[width=0.45\textwidth]{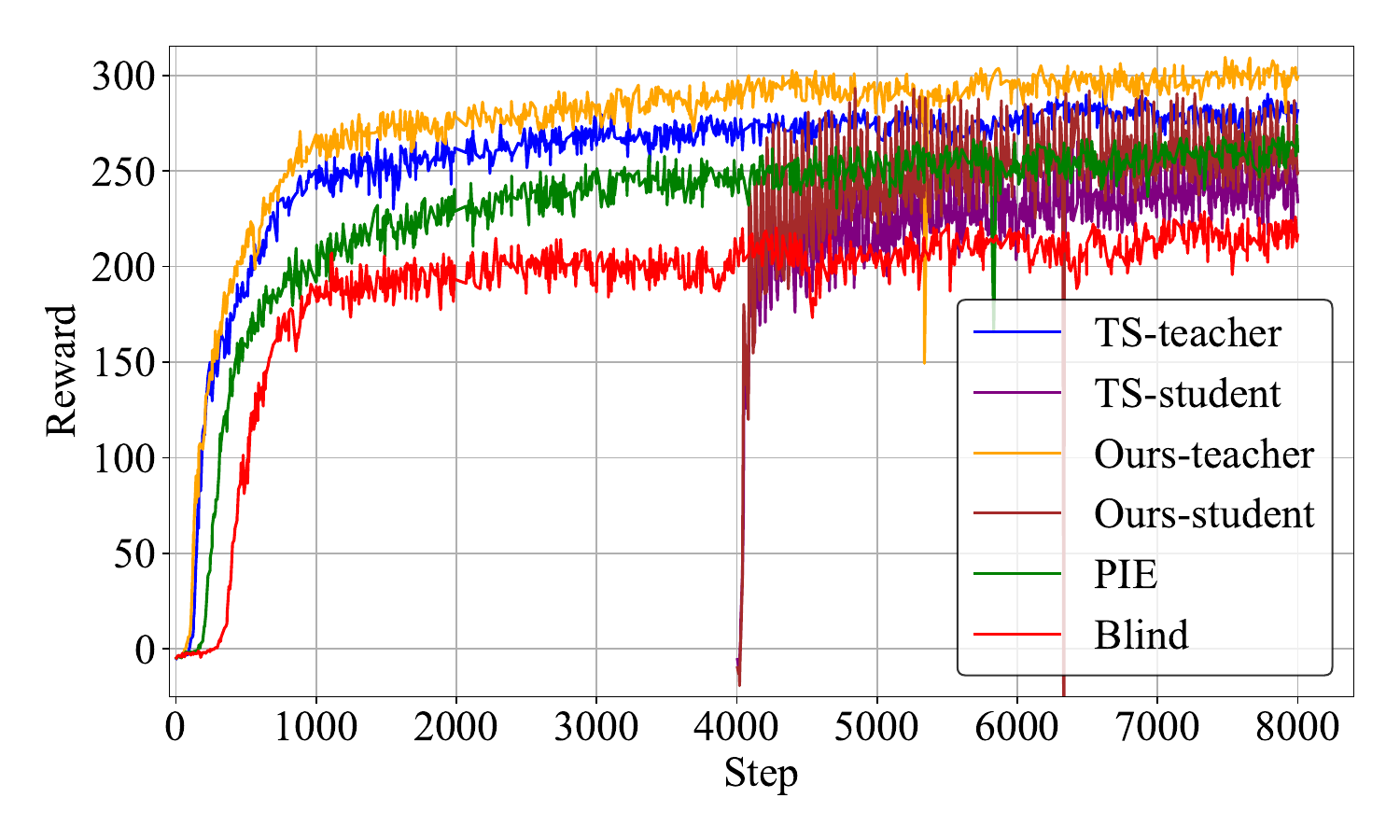}  
    \caption{The reward curves during the training process of our method and several other approaches.}
    \label{fig:reward}
\end{figure}

\begin{figure}[htbp]
    \centering
    \includegraphics[width=0.45\textwidth]{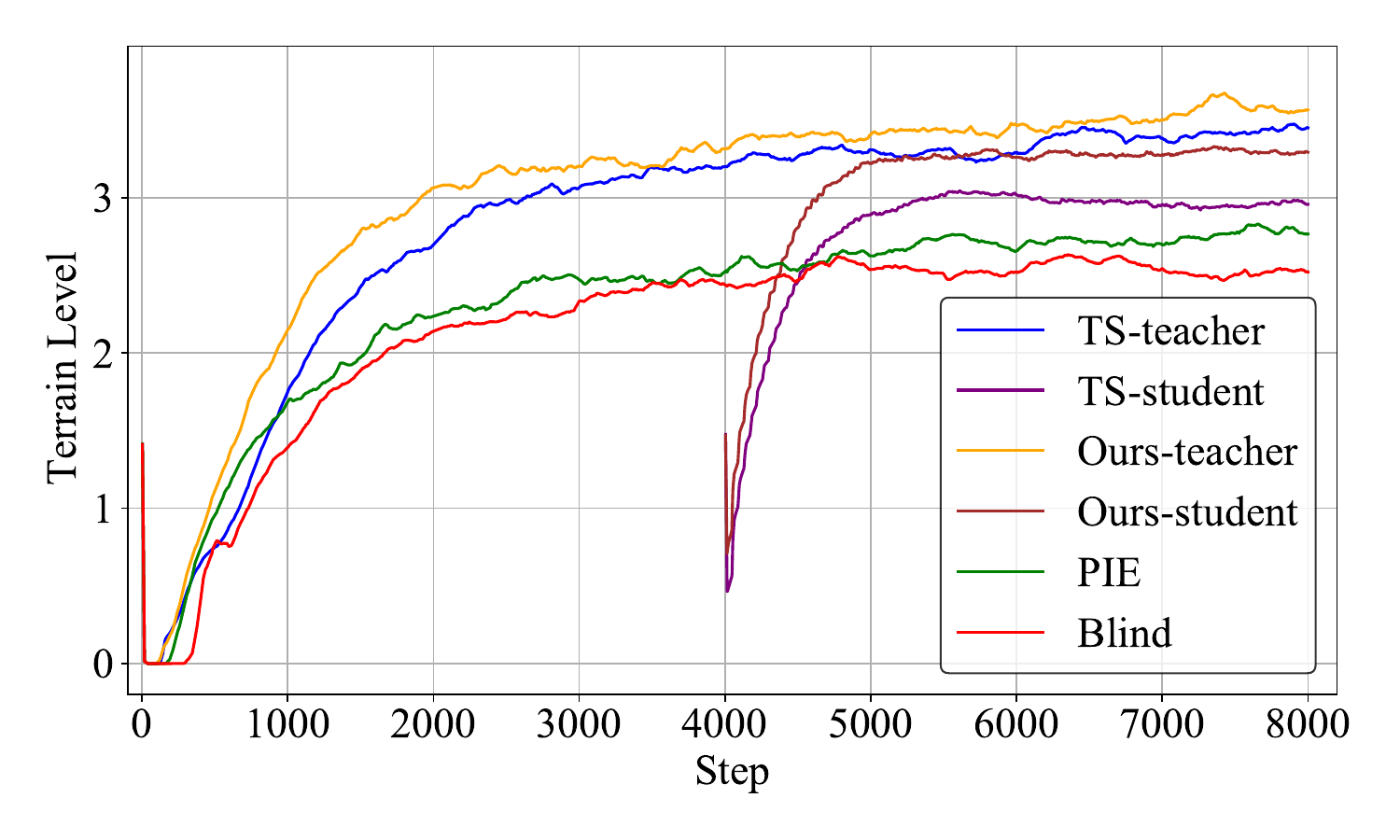}  
    \caption{The terrain level curves during the training process of our method and several other approaches.}
    \label{fig:level}
\end{figure}

We selected the average reward and terrain level as evaluation metrics during the training process. The average reward serves as an indicator of the quality of the training task, while the terrain level assesses the robot's ability to overcome various terrains. The average reward and terrain level curves during the training processes of our method and several other approaches are presented in Fig. \ref{fig:reward} and Fig. \ref{fig:level}, respectively. 

Fig. \ref{fig:reward} illustrates that, during the training process, the reward of our teacher policy exceeds that of the TS-teacher. The reward of the student policy is influenced by the performance of the teacher policy, resulting in a higher reward for our student policy compared to TS-student. In contrast, PIE focuses more on the velocity term due to its prediction of variables such as speed, thereby assigning higher weight to the velocity component in the reward function. Consequently, the final reward of our method closely approximates that of PIE. Fig. \ref{fig:level} illustrates that the terrain level of our student policy surpasses that of TS-student. Although PIE incorporates terrain prediction, it still struggles to effectively overcome challenging terrains. Consequently, the terrain level of our student policy exceeds that of PIE, while PIE performs better than the blind policy.
\begin{figure}[htbp]
    \centering
    \includegraphics[width=0.45\textwidth]{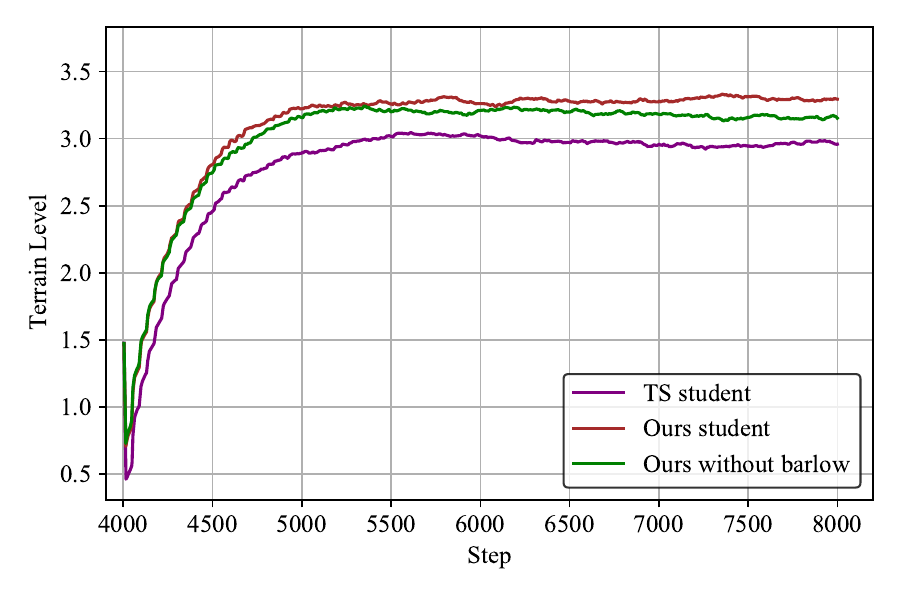}  
    \caption{The terrain level curves during the training process of the TS policy, incorporating the Barlow Twins method, and our proposed method.}
    \label{fig:barlow_level}
\end{figure}

To validate the effect of the alignment module (Barlow Twins) on the student policy, we compared the terrain level of the TS strategy, the TS strategy with the alignment module, and our proposed method. The results, as shown in Fig. \ref{fig:barlow_level}, demonstrate that the terrain level of the strategy with the alignment module exceeds that of the original TS strategy, yet still falls short of the performance achieved by our method.
\begin{figure}[htbp]
    \centering
    \includegraphics[width=0.5\textwidth]{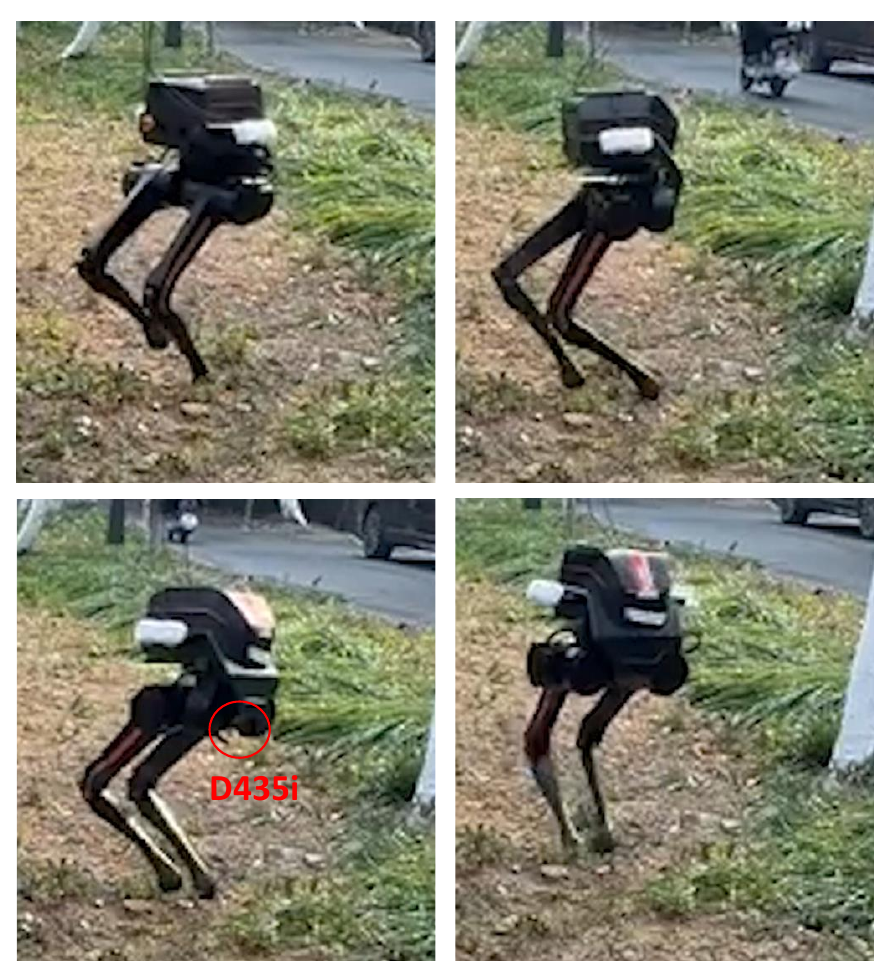}  
    \caption{The bipedal robot maneuvers and turns on grassland.}
    \label{fig:grass}
\end{figure}

\begin{table}[htbp]
\centering
\caption{Evaluate Result}
\begin{tabular}{cccc}
\toprule  
\textbf{Method} & Vel. Error &  Height Error & Survival Time \\
\midrule  
Blind& 0.929 & 0.076 & 84.675 \\
PIE&0.546 & 0.084 & 87.111 \\
TS& 0.673 & 0.097 & 88.601 \\
Ours& \textbf{0.535} & \textbf{0.071} & 88.485 \\
Ours w/o barlow& 0.554 & 0.079 & \textbf{88.652} \\
Ours w/o Moe& 0.593 & 0.093 & 87.011 \\
\bottomrule  
\end{tabular}
\label{tab:res}
\end{table}
In the simulation, we deployed 128 robots across several terrains with fixed proportions to evaluate the performance of different strategies in terms of speed tracking, average survival time within 100 seconds, and height tracking error, which are considered fundamental performance metrics. The results are presented in Table \ref{tab:res}. In the simulation experiments, our framework outperforms both the teacher-student policy and the PIE strategy in terms of basic user command tracking and survival time. In the case of w/o Moes, our strategy exhibits slightly lower velocity tracking performance compared to PIE, which is attributed to PIE's real-time estimation of the robot's state. However, in the case of w/o Barlow Twins, our strategy shows a significant improvement over the TS policy, although its velocity tracking error still slightly exceeds that of the PIE strategy.
\begin{figure*}[t]
    \centering
    \includegraphics[width=\textwidth]{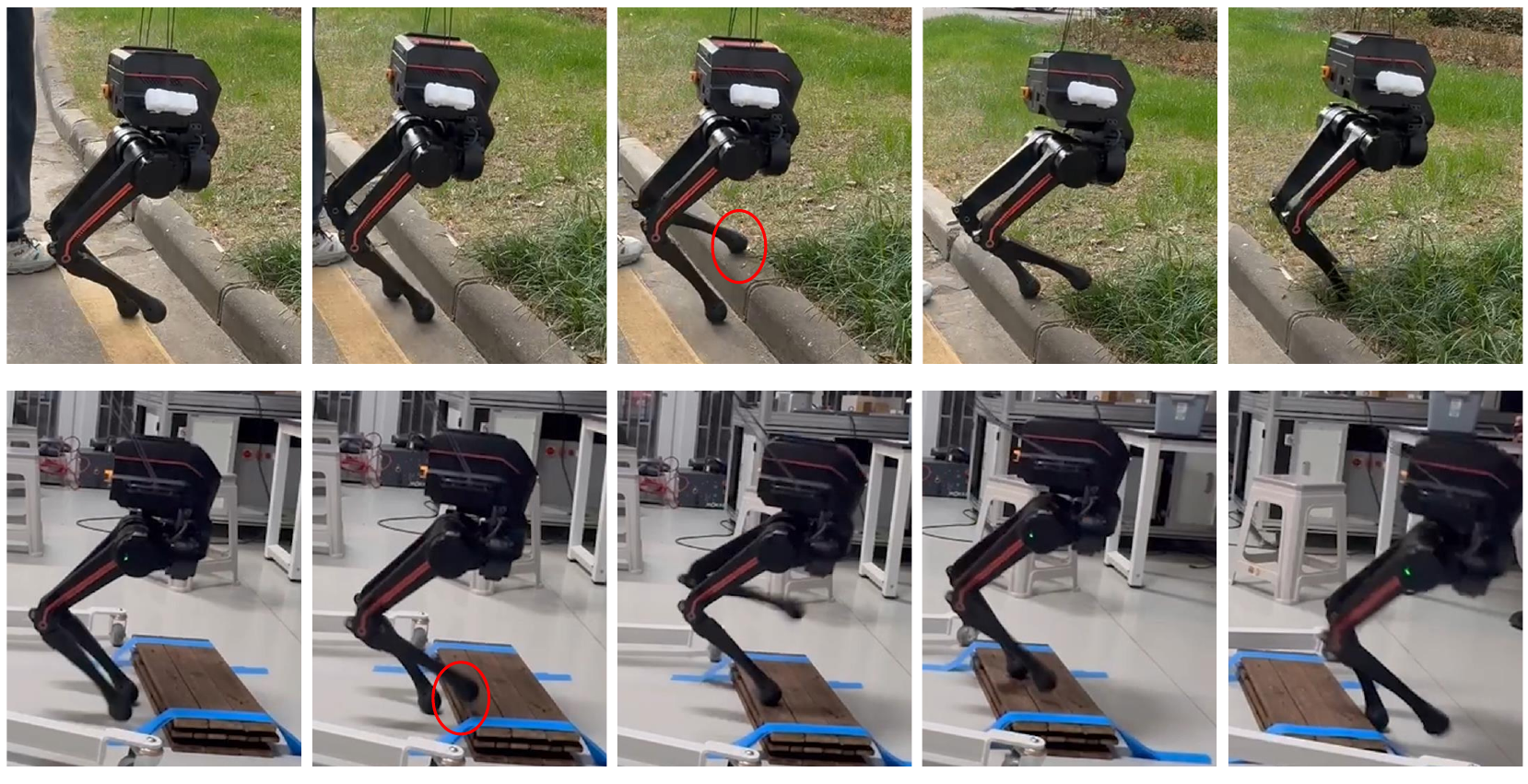}  
    \caption{The bipedal robot adjusts its gait to overcome obstacles upon detecting them. The red circle in the figure indicates the robot autonomously raising its foot height to traverse the terrain.}
    \label{fig:action_swq}
\end{figure*}
\subsection{Real-World Experiments}
We deployed our strategy on the Limx Dynamics P1 robot, which features a 6-DOF body and is equipped with an IMU and a D435i sensor. We tested our strategy across various terrains.

As shown in Fig. \ref{fig:grass}, we deployed our strategy on grassland. Despite the soft grass not providing the same stable friction and elasticity as a flat surface, the bipedal robot is still able to achieve stable locomotion and turning.

\begin{figure}[htbp]
    \centering
    \includegraphics[width=0.5\textwidth]{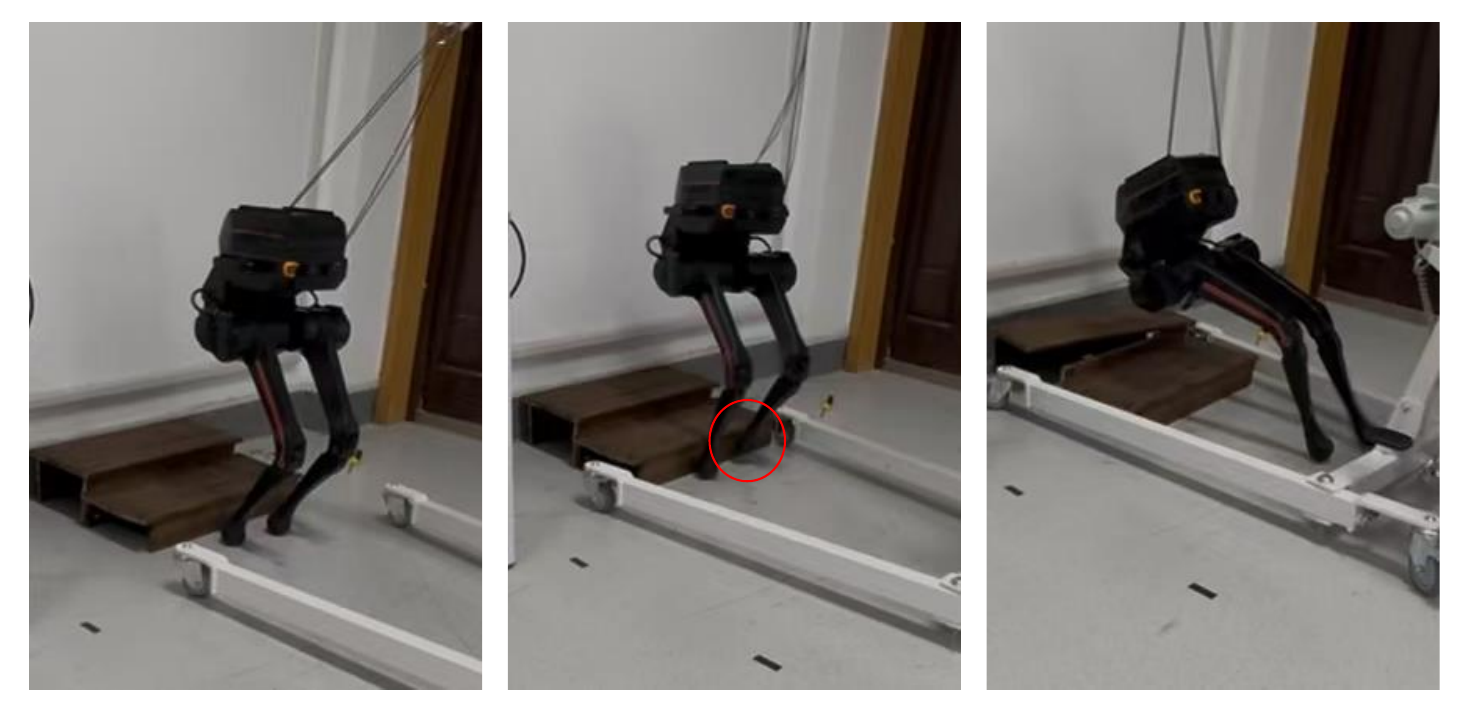}  
    \caption{The robot does not autonomously raise its foot height and falls while blind walking. The red circle in the figure indicates the point where the foot makes contact with the vertical portion of the step.}
    \label{fig:novi}
\end{figure}

To validate the robot's ability to recognize and overcome obstacles, we evaluated its capacity to traverse stairs and surfaces elevated above the current ground level. As shown in Fig. \ref{fig:action_swq}, when the robot detects an obstacle that can be crossed, it increases the foot's lift height to ensure stable clearance. After overcoming the obstacle, the robot lowers the foot height to maintain its stability. This demonstrates that our strategy, supported by vision, enables the robot to effectively navigate over obstacles. As shown in Fig. \ref{fig:novi}, the strategy that does not use camera depth images as input is highly prone to falling due to unintended contact with unknown terrain.
\section{CONCLUSION AND FUTURE WORKS}
We propose a novel vision-based reinforcement learning strategy that combines the mixture of experts model, contrastive learning, and teacher-student models, yielding slightly better performance than the teacher-student model alone. We validated the effectiveness of this approach through both simulation comparisons and real-world deployment, demonstrating its capability to traverse various terrains. However, the inherent instability of the chosen bipedal platform, such as difficulty in maintaining a stable speed, posed significant challenges to our experiments. Furthermore, the current work is limited to control and does not address trajectory planning. Therefore, future work will focus on enhancing the stability of the bipedal robot and developing dynamic path planning strategies.

\bibliographystyle{IEEEtran}  
\bibliography{references}  

\end{document}